\documentclass[10pt,twocolumn,letterpaper]{article}

\usepackage{cvpr}
\usepackage{times}
\usepackage{epsfig}
\usepackage{graphicx}
\usepackage{amsmath}
\usepackage{amssymb}
\usepackage{hyperref}
\usepackage{booktabs,multirow, multicol}

\graphicspath{./figures}

\newcommand{\hfr}{\textit{HFR}{} }

\usepackage{xcolor}
\usepackage{colortbl}

\definecolor{Gray}{rgb}{0.9, 0.9, 0.9}

\newcommand{\ssmbfull}{Switch Style Modulation Blocks{} }
\newcommand{\ssmb}{\textit{SSMB}{} }

\begin{document}

\title{\LARGE \bf Modality Agnostic Heterogeneous Face Recognition with Switch Style Modulators}

\author{Anjith George and S\'ebastien Marcel \\
Idiap Research Institute \\
Rue Marconi 19, CH - 1920, Martigny, Switzerland \\
{\tt\small  \{anjith.george, sebastien.marcel\}@idiap.ch  }
}

\maketitle
\thispagestyle{empty}

\begin{abstract}

Heterogeneous Face Recognition (HFR) systems aim to enhance the capability of face recognition in challenging cross-modal authentication scenarios. However, the significant domain gap between the source and target modalities poses a considerable challenge for cross-domain matching. Existing literature primarily focuses on developing HFR approaches for specific pairs of face modalities, necessitating the explicit training of models for each source-target combination. In this work, we introduce a novel framework designed to train a modality-agnostic HFR method capable of handling multiple modalities during inference, all without explicit knowledge of the target modality labels. We achieve this by implementing a computationally efficient automatic routing mechanism called Switch Style Modulation Blocks (SSMB) that trains various domain expert modulators which transform the feature maps adaptively reducing the domain gap. Our proposed SSMB can be trained end-to-end and seamlessly integrated into pre-trained face recognition models, transforming them into modality-agnostic HFR models. We have performed extensive evaluations on HFR benchmark datasets to demonstrate its effectiveness. The source code and protocols will be made publicly available.

\end{abstract}

\section{Introduction}

Face recognition (FR) technology has become a popular choice for access control due to its efficiency and user-friendliness, often achieving human-level performance \cite{learned2016labeled}. However, FR systems typically operate within a homogeneous domain, using RGB camera-captured facial images for both enrollment and matching. Nevertheless, there are situations where matching in a heterogeneous setting, incorporating different sensing modalities is advantageous. For instance, NIR cameras offer improved performance in varying lighting conditions and spoofing resilience \cite{li2007illumination,george2022comprehensive}, yet developing an NIR-based FR system requires large-scale annotated data which is often difficult to get.  Heterogeneous Face Recognition (\hfr\!) systems address this by enabling cross-domain matching without the requirement of enrolling with the target modality, proving valuable in challenging conditions. Heterogeneous face recognition (HFR) refers to the process of recognizing faces across different sensor modalities or image domains. For example, an HFR system can perform biometric matching between a gallery of identities enrolled using RGB images with images captured from another sensor like NIR CCTV cameras, thermal cameras, low-resolution video, and so on. In essence, HFR systems enable cross-domain matching without the requirement of enrollment with each of the domains. Such systems extend FR's utility to scenarios like low-light or long-range recognition using various imaging modalities. However, the domain gap between modalities makes this challenging \cite{he2018wasserstein}. Further, the limited availability of paired data makes this even more challenging to train these models.

Most of the existing works in literature try to reduce the domain gap using a pair of source target modality combinations. Several methods have been reported in the literature to address the heterogeneous face recognition task. Invariant Feature-based approaches like Difference of Gaussian (DoG) and scale-invariant feature transform (SIFT) have been proposed to mitigate the domain gap in \hfr\! \cite{liao2009heterogeneous, klare2010matching}. Common-space projection methods represent another category, focusing on learning mappings that project diverse facial modalities into a shared subspace to bridge the domain gap \cite{kan2015multi, he2017learning}. Recent advancements in \hfr\! methods, such as DVG and generative adversarial network (GAN)-based synthesis, offer high-quality image generation, albeit at increased computational costs, limiting practical utility \cite{fu2021dvg, zhang2017generative}. A novel approach introduced in \cite{de2018heterogeneous}, termed Domain-Specific Units (DSU), proposes that CNNs' high-level features trained on visible spectrum data can encode images from other modalities, implying their domain-independence. In contrast, Prepended Domain Transformers (PDT) \cite{george2022prepended} augments a pre-trained FR network with a dedicated module for the target modality, transforming it into an \hfr\! network. In a different approach, outlined in \cite{george2023bridging}, different modalities are treated as distinct styles and cross-modal matching is achieved through the modulation of feature maps in the target modality using conditional adaptive instance modulation (CAIM).

In most of these works, HFR is addressed by using a pair of modalities, a reference modality (often RGB images), and a target modality such as thermal, NIR, SWIR, and so on. Training with a pair of modalities requires one to know what modalities to expect at test time, and the HFR model needs to be trained for that specific pair of modalities. Also, there exist separate computational paths for two modalities, this is problematic in scenarios where this information is not available,
for instance in the case of low-resolution and high-resolution images; passing high-resolution images in the low-resolution path may degrade performance.
 This limitation restricts their scalability as they need to be trained for each pair of modalities, which is impractical when dealing with a wide range of possible modalities. This becomes rather challenging as HFR is already a data-limited problem.

Knowing that there is shared information due to the face structure across modalities it could be possible to develop a model which is modality-agnostic and universal for modalities across the board.  In this work, we propose \ssmbfull (SSMB) which introduces modality-agnostic HFR by automatically routing input images, eliminating the need to know the modality during testing. This automatic routing mechanism enhances computational efficiency and broadens the applicability of the model to various face modalities using a single universal model. Moreover, this approach is data-efficient, requiring only a limited number of samples from different modalities for training. By exploiting shared information across face representation in various modalities, the \ssmb approach addresses key challenges in HFR, making it a versatile and efficient solution for cross-modal face recognition.

The proposed approach is sensor-agnostic and capable of processing input from diverse sources, including CCTV cameras, mobile phones, infrared sensors, and hand-drawn sketches. This versatility is vital in applications like surveillance, where input varies widely. Even in scenarios with low-quality images, such as old databases, sensor-agnostic HFR systems can be useful for recognition tasks. They can also seamlessly integrate with emerging technologies like augmented reality, virtual reality, and robotics, which employ various sensors to capture facial data.

The main contributions of this work can be summarized as follows:

  \begin{itemize}
    \item We introduce a modality-agnostic Heterogeneous Face Recognition (HFR) approach, expanding the applicability of HFR beyond pairs of specific modalities.
    \item The proposed approach is data and computationally efficient.
    \item We introduce new protocols to evaluate the performance using the MCXFace dataset.
    \item We have performed a comprehensive evaluation of the proposed approach across various protocols and datasets demonstrating its effectiveness.
  \end{itemize}
  
Finally, the protocols and source codes will be made available publicly \footnote{\url{https://gitlab.idiap.ch/bob/bob.paper.ijcb2024_moe_hfr}}.

\section{Related works}
\label{sec:relatedworks}

Heterogeneous Face Recognition (HFR) methods aim to match faces from images taken with different sensing technologies. However, the domain gap between these modalities can reduce the effectiveness of face recognition systems in directly comparing multi-modal images. Thus, it's essential for HFR approaches to bridge this modality gap. This section reviews recent approaches proposed to tackle the domain gap.

\textbf{Invariant feature-based methods:} In Heterogeneous Face Recognition (HFR), various strategies have been proposed in the literature to extract invariant features across different sensing modalities. Liao et al. \cite{liao2009heterogeneous} presented a method combining the Difference of Gaussian (DoG) filters with multi-scale block Local Binary Patterns (MB-LBP) for invariant feature extraction. Klare et al. \cite{klare2010matching} employed Local Feature-based Discriminant Analysis (LFDA), incorporating Scale-Invariant Feature Transform (SIFT) and Multi-Scale Local Binary Pattern (MLBP) as descriptors. Zhang et al. \cite{zhang2011coupled} introduced Coupled Information-Theoretic Encoding (CITE), aiming to maximize mutual information across modalities in quantized feature spaces. The use of Convolutional Neural Networks (CNNs) has been explored for HFR, demonstrating its effectiveness in this domain \cite{he2017learning,he2018wasserstein}. Roy et al. \cite{roy2018novel} developed a Local Maximum Quotient (LMQ) to identify invariant features in cross-modal facial images. In \cite{liu2018composite}, a method for composite sketch recognition was proposed, using Scale-Invariant Feature Transform (SIFT) and Histogram of Oriented Gradient (HOG) for feature extraction and integrating these at the score level with a linear function to combine facial components.

\textbf{Common-space projection methods:} Common-space projection methods aim to project facial images from diverse modalities into a single subspace, reducing the domain gap \cite{kan2015multi,he2017learning}. Lin and Tang \cite{lin2006inter} introduced common discriminant feature extraction for cross-modal image alignment. Yi et al. \cite{yi2007face} applied Canonical Correlation Analysis (CCA) to NIR and VIS images. Lei et al. \cite{lei2009coupled,lei2012coupled} developed regression-based mappings for reducing the modality gap. Sharma and Jacobs \cite{sharma2011bypassing} used Partial Least Squares (PLS) for linear mapping across modalities. Klare and Jain \cite{klare2012heterogeneous} projected face images onto a linear discriminant analysis subspace using prototype similarities. De Freitas et al. \cite{de2018heterogeneous} showed that CNN high-level features are domain-agnostic, utilizing Domain-Specific Units (DSUs) for domain gap minimization. Liu et al. \cite{liu2020coupled} introduced Coupled Attribute Learning for HFR (CAL-HFR) and Coupled Attribute Guided Triplet Loss (CAGTL) for shared space mapping without manual labeling. Recently, Liu et al. \cite{liu2023modality} proposed a semi-supervised method, Modality-Agnostic Augmented Multi-Collaboration representation for HFR (MAMCO-HFR), leveraging network interactions for discriminative information extraction and introducing a technique for adversarial perturbation-based feature mapping.

\textbf{Synthesis based methods:} Synthesis-based methods in Heterogeneous Face Recognition (HFR) generate synthetic images in the source domain from the target modality, enabling the use of conventional face recognition networks \cite{tang2003face,fu2021dvg}. Wang et al. \cite{wang2008face} developed a patch-based synthetic approach using Multi-scale Markov Random Fields. Liu et al. \cite{liu2005nonlinear} utilized Locally Linear Embedding (LLE) for pixel-wise matching between VIS images and sketches. CycleGAN's introduction for unpaired image translation \cite{zhuUnpairedImagetoImageTranslation2017} has facilitated transforming target domain images to the source domain \cite{baeNonvisualVisualTranslation2020}. Zhang et al. \cite{zhang2017generative} employed Generative Adversarial Networks (GANs) for generating photo-realistic VIS images from thermal images via GAN-based Visible Face Synthesis (GAN-VFS). The Dual Variational Generation (DVG-Face) framework \cite{fu2021dvg} leverages GANs for VIS image synthesis, showing good performance in \hfr benchmarks. Liu et al. \cite{liu2021heterogeneous} introduced Heterogeneous Face Interpretable Disentangled Representation (HFIDR) for latent identity information extraction and cross-modality synthesis. Luo et al. \cite{luo2022memory} introduced a Memory-Modulated Transformer Network (MMTN) which addresses HFR as an unsupervised, reference-based generation problem, combining prototypical style patterns and style blending. Recently, George et al. \cite{george2022prepended} proposed Prepended Domain Transformers (PDT), a module prepended to a pre-trained FR network for cross-domain feature alignment, eliminating the need for explicit source domain image generation.

\textbf{Limitations of HFR approaches:} Many of the recent HFR methods adopt a synthesis-based approach. However, this requires the synthesis of high-fidelity RGB images from the target modality and is computationally expensive. Further, most of the methods are trained explicitly for a pair of modalities, and they need to be trained for every pair of modalities separately. This makes the problem more challenging as the HFR problem is already data-limited.
\section{Proposed method}

\label{sec:proposed}

\begin{figure*}[t!]
  \centering
      \includegraphics[width=0.79\linewidth]{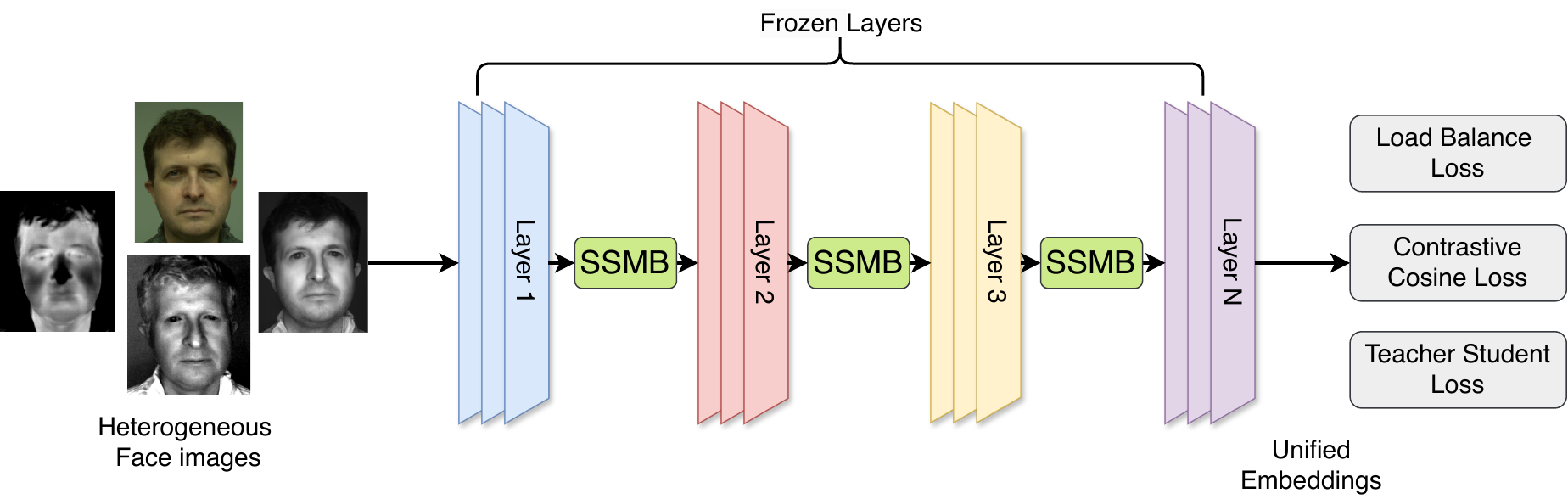}
  \caption{Architecture of the proposed framework, the \ssmbfull adaptively routes the samples (\ssmb) such that identities for multiple face modalities result in a modality agnostic unified embedding. The embeddings can be used for homogeneous or heterogeneous matching between any pair of modalities.}
  \label{fig:framework}
\end{figure*}

Most of the approaches proposed in \hfr specifically focus on cross-domain matching between a pair of modalities such as Visible-Thermal, Visible-Near Infrared, and so on. This requires the \hfr models to be trained for each pair of modalities explicitly. This also reduces the opportunities to capture generic features and cross-modal consistencies in an already data-scarce domain like HFR. Further, most of the methods available in the literature assume that the label for modality is already known and uses an asymmetric pipeline for processing the data-- meaning the dataflow is different for source and target modalities. 

In this work, we go beyond the traditional notion of heterogeneous face recognition and design a framework that can handle multiple face imaging modalities at the same time without needing explicit labels for different modalities. We propose to use an efficient mixture of expert-type routing mechanism inside our framework which can automatically infer the path for the specific input modality. The proposed framework could be used for matching between multiple modalities (not just reference and target) without any labels on the modalities. This makes our approach a universal way of doing cross-modality face matching.

\subsection{Definition of HFR}

In Heterogeneous Face Recognition (\hfr), we consider a domain $\mathcal{D}$ containing samples $X \in \mathbb{R}^d$ with a marginal distribution $P(X)$ and dimensionality $d$. A face recognition (FR) system, $\mathcal{T}^{fr}$, aims to associate these samples with labels $Y$ through a model parameterized by $\Theta$, described by the conditional probability $P(Y|X,\Theta)$. During training, this model learns from a dataset of faces $X={x_1, x_2, ..., x_n}$ and corresponding identity labels $Y={y_1, y_2, ..., y_n}$ via supervised learning.

In \hfr, we differentiate between a source domain $\mathcal{D}^s = {X^s, P(X^s)}$ and a target domain $\mathcal{D}^t = {X^t, P(X^t)}$, both linked by the label space $Y$. The \hfr problem involves finding $\hat{\Theta}$ to align conditional probabilities across domains, i.e., $P(Y|X^s, \Theta) = P(Y|X^t, \hat{\Theta})$.

In our framework, we go beyond \hfr and propose a Modality Agnostic Heterogeneous Face Recognition (MAHFR) which can include multiple modalities of images

$\mathcal{D}^t_{j} = {X^t_{j}, P(X^t_{j})}$, for different modalities $j$ such as thermal,  near infrared and so on.

And we formulate the MAHFR problem as estimating $\hat{\Theta}$ to align conditional probabilities across domains, i.e., $P(Y|X^s, \Theta) = P(Y|X^t_{j}, \hat{\Theta})$.

\subsection{Overall architecture}

The architecture of the framework proposed is shown in Fig. \ref{fig:framework}. We start with a pre-trained face recognition model and insert a trainable module between the frozen layers. The objective is that this architecture produces similar embeddings across different face modalities for a specific subject, such that embedding can be used for biometric matching with a cosine similarity metric. The unified embedding obtained is modality agnostic and can be used for both homogeneous and heterogeneous experiments. These \ssmb modules need to be adaptive in the sense that the operation performed by these blocks should change based on the input sample.
The details of the components of this framework are given in the following sub-sections.

\subsection{Switch Style Modulation for HFR}

The primary challenge in cross-modal face recognition lies in bridging the domain gap between different modalities. Conditional Adaptive Instance Modulation (CAIM), as proposed in the works by George et al. \cite{george2023bridging, george2024modalities}, addresses this issue by integrating instance modulation modules within the layers of a pre-trained Face Recognition (FR) network. These modules, placed between the network's frozen layers, are controlled by an external gate, leading to an asymmetric pipeline that necessitates modality labels at the inference stage for selecting the computational path to use.

\begin{figure}[t!]
  \centering
  \includegraphics[width=0.89\linewidth]{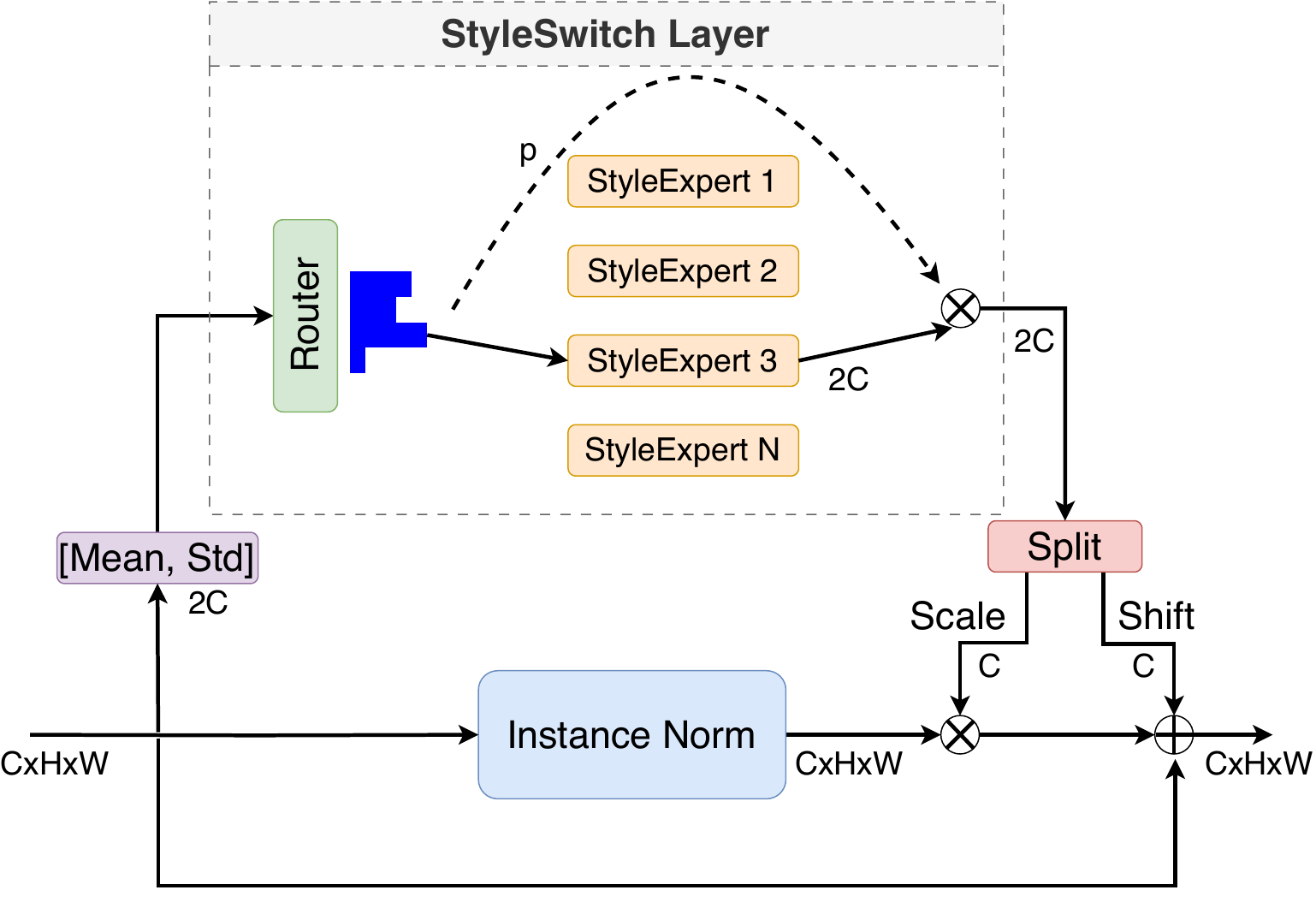}
  \caption{Architecture of the Switch Style modulation block (SSMB). The inputs and outputs of this module are feature-maps of dimension $C\times H \times W$.}
  \label{fig:ssmb_arch}
\end{figure}

Building on this, our approach introduces an adaptive mechanism that automatically routes samples from different modalities, eliminating the dependency on an external gating mechanism. This allows for the development of a modality-agnostic representation within a unified framework making HFR more efficient. The architecture of our proposed modulation block, denoted as the \ssmb, is illustrated in Fig. \ref{fig:ssmb_arch}. We simplify the design of the modulators with a single fully connected layer. We use an adaptive routing mechanism based on a Mixture of experts framework for sample-based routing.

Mixture of experts (MoE),\cite{jacobs1991adaptive,jordan1994hierarchical,shazeer2017outrageously} enables neural networks to have sample dependant parameters thereby enhancing the abilities of the base model. In the context of large language models (LLMs), MoE \cite{shazeer2017outrageously} routes a token representation $x$ dynamically to top-$k$ experts, from a pool of $\{E_i(x)\}_{i=1}^N$ of $N$ experts. A router module computes logits over the number of experts available with a softmax operation and top-$k$ gate values and experts are selected. The output of the MoE layer can be computed as the weighted combination of the gate values and expert outputs.

\begin{equation}\label{eqn: moe_layer}
    y = \sum_{i \in \mathcal{T}} p_i(x) E_i(x). 
\end{equation}
where, $\mathcal{T}$ is the set top-$k$ experts selected, and $E_i(x)$ the output of $i^{th}$ expert.

Switch Transformers, as proposed by Fedus et al. \cite{fedus2022switch}, introduced a concept of sparsely activated expert models known as switch layers. In this design, for each input sample, the network activates only a subset of its network blocks, thereby maintaining a constant computational requirement. Contrary to other works \cite{shazeer2017outrageously, ramachandran2018diversity} where the routing is performed to $k>1$ expert, switch transformer route only to a single expert, referred to as switch routing. The gate value $p_i(x)$ in Equation \ref{eqn: moe_layer} allows the differentiability of the router module. They also add a load balance loss as an auxiliary loss to balance the routing of samples to different experts.

In this work, rather than establishing distinct models for various segments of the data, we solely utilize modulators within the Mixture-of-Expert (MoE) framework. Specifically, we diverge from the traditional MoE approach and construct a \textit{StyleSwitch} Layer, signifying that modulation is exclusively conducted through routers.  We adopt the switch layers where only one expert is activated for a specific sample. This approach allows the model to select sample-dependent modulation parameters at the same time keeps the computational burden small. 

For a feature map $\mathbf{F}$, of dimension $C \times H \times W$, we first compute the mean and std deviation of the feature map along channel dimension which are concatenated as the input for the \textit{StyleSwitch} layer.

\begin{equation}
    RouterInput=[\boldsymbol{\mu}(\mathbf{F}), \boldsymbol{\sigma}(\mathbf{F})]
\end{equation}
The router is implemented as a single fully connected layer with $N$ output nodes and a softmax layer, where $N$ is the number of experts.

To estimate the modulation parameters -- scale and shift, we use a \textit{StyleExpert} layer, which is implemented as a single fully connected layer. The weight matrix of the \textit{StyleExpert} layer is initialized as an identity matrix so that the training is stable. The \textit{StyleSwitch} layer performs the routing and obtains the final output. Only one expert is activated due to the top-1 routing mechanism. The output of the \textit{StyleSwitch} module has the dimension $2C$, where $C$ is the number of input channels, we split this to get the new modulation parameters.

\begin{equation}
    \mathbf{\sigma_{s}},\mathbf{\mu_{s}} = \mathbf{split}(StyleSwitchOutput)
\end{equation}

Finally, we scale and shift the instance normalized feature-maps using the output of the \textit{StyleSwitch} layer. We further add a residual connection for stable training. Overall the operation of the \ssmb can be represented as:

\begin{equation}
  \mathrm{SSMB}(\mathbf{F}) = \frac{1}{2} (\mathbf{\sigma_{s}} \left(\frac{\mathbf{F} - \boldsymbol{\mu}(\mathbf{F})}{\boldsymbol{\sigma}(\mathbf{F})}\right) + \mathbf{\mu_{s}} + \mathbf{F})
\end{equation}

\subsection{Training paradigm}

To train this module in an end-to-end fashion we use contrastive loss together with a teacher-student framework similar to the one proposed in \cite{george2024heterogeneous}. We will denote the network we are training (Fig. \ref{fig:framework}) as a student network in this context. For training, we need the labels for the reference modality samples (VIS) during training, but explicit labels for non-source modalities are unnecessary as they are grouped as a a single target modality. We create pairs of samples from the source modality and target modality, note that this target modality includes any modality different from the source modality. Pairs with the same identity have a label 1 and pairs with dissimilar identity has a label 0. We use three loss functions in our training.

First, we use a cosine contrastive loss function, denoted by $\mathcal{L}_{C}$ to align the embeddings of different modalities together, such that the same identity clusters together and different identities are pushed apart:

\begin{equation}
  \resizebox{.99\linewidth}{!}{$
  \begin{aligned}
  \mathcal{L}_{C}(e_{S_{s_{i}}}, e_{S_{t_{i}}}, y_i) = & (1 - y_i) \cdot \max\left(0, \frac{e_{S_{s_{i}}} \cdot e_{S_{t_{i}}}}{\left\| e_{S_{s_{i}}} \right\|_2 \left\| e_{S_{t_{i}}} \right\|_2} - m\right) \\
  & +  y_i \cdot \left(1 - \frac{e_{S_{s_{i}}} \cdot e_{S_{t_{i}}}}{\left\| e_{S_{s_{i}}} \right\|_2 \left\| e_{S_{t_{i}}} \right\|_2}\right)
  \end{aligned}
  $}
  \end{equation}

Where $m$ denotes the margin, $e_{S_{s_{i}}}$ and $e_{S_{t_{i}}}$ denote the embedding for the source modality and target modality from the student network (under training).

Second, due to the limited size of the HFR datasets we use teacher-student supervision using an identity loss computed from the same pretrained face recognition backbone.  It is to be noted that this is applied only to the images in the reference modality as the teacher model is to be used for the visible modality \cite{george2024heterogeneous}.

We use the following Teacher Student Identity loss $\mathcal{L}_{TSI}$:
\begin{equation}
\mathcal{L}_{TSI}(e_{T_{s_{i}}}, e_{S_{s_{i}}}) = \left(1 - \frac{e_{T_{s_{i}}} \cdot e_{S_{s_{i}}}}{\| e_{T_{s_{i}}} \|_2 \| e_{S_{s_{i}}} \|_2}\right)
\end{equation}

where $e_{T_{s_{i}}}$ denotes the embedding for the source modality image from the pre-trained teacher. 

Third, to balance the utilization of different experts we also use a load balancing loss  $\mathcal{L}_{b}$ as reported in \cite{fedus2022switch}.

The overall loss function to optimize can be written as:

\begin{align}
  \mathcal{L}(e_{S_{s_{i}}}, e_{S_{t_{i}}},e_{T_{s_{i}}}, y_i) = & (1- \gamma) \cdot \mathcal{L}_{C}(e_{S_{s_{i}}}, e_{S_{t_{i}}}, y_i)  \\
  &+ \gamma \cdot \mathcal{L}_{TSI}(e_{T_{s_{i}}}, e_{S_{s_{i}}}) + \alpha \cdot \mathcal{L}_{b}
  \end{align}

where $\gamma$ is a hyper-parameter that influences the relative importance of contrastive loss and teacher-student supervision. Parameter $\alpha$ controls the contribution of load balancing loss in the overall loss.

\subsection{Face Recognition backbone}
For reproducibility, we utilized the publicly available pre-trained \textit{Iresnet100} face recognition model from Insightface \cite{insightface}, trained on the MS-Celeb-1M dataset featuring over 70,000 identities \footnote{\url{http://trillionpairs.deepglint.com/data}}. This model processes three-channel images of $112 \times 112$ pixels. Faces are aligned and cropped to match eye center coordinates with specific reference points before analysis. For single-channel inputs (e.g., NIR or thermal images), the channel is replicated across all three channels without modifying the network structure following \cite{george2024modalities}.
\subsection{Implementation details}
The proposed approach is trained end-to-end, supervised by three distinct loss functions as described in the previous section. For images in the visible spectrum, we compute the identity loss using both the pre-trained teacher backbone and the student model equipped with \ssmb layers. Additionally, a load balancing loss and a contrastive cosine loss are incorporated into the optimization process during training. The implementation framework utilizes PyTorch and integrates with the Bob library \cite{bob2017,bob2012}\footnote{\url{https://www.idiap.ch/software/bob/}}. We employ the Adam Optimizer with an initial learning rate of $0.0001$, trained for 50 epochs with a batch size of 48. The margin parameter $m$ is set to zero, and the hyper-parameter $\gamma$ is set to 0.50. The $\alpha$ parameter is set as 0.01 in all the experiments.
We initialize the network layers of our student model from the weights of the pre-trained face recognition network. We use three \ssmb blocks in the initial layers of the student model. The weights of the \textit{StyleExperts} layer are initialized as an identity matrix so that it produces the same embedding at the beginning of training. Throughout the training phase, only the \ssmb layers are adapted, while the remainder of the layers remain frozen. During the inference phase, the teacher network is removed, allowing the same student network to process both source and target domain samples without the need for modality labels. This enables the generation of modality-agnostic facial embeddings. Verification and identification of experiments are then performed using cosine distance metric on these unified embeddings.

\section{Experiments}
\label{sec:experiments}

\begin{table*}[htb!]
  \centering
  \caption{Experimental results on VIS-UNIVERSAL protocol of the MCXFace dataset-- aggregated performance.}
  \label{tab:mcxface_combined}
  \resizebox{0.7\textwidth}{!}{

\begin{tabular}{cccccc}
\toprule
\textbf{Modality} &             \textbf{AUC} &           \textbf{EER} &              \textbf{Rank-1} &         \textbf{VR@FAR=0.1$\%$} &            \textbf{VR@FAR=1$\%$} \\
\midrule
DSU \cite{george2022prepended} &  95.57$\pm$0.80 & 10.24$\pm$0.88 &  84.21$\pm$0.94 &  67.89$\pm$1.05 &  78.13$\pm$1.20 \\
PDT \cite{george2022prepended} & 96.16$\pm$1.60 &  9.60$\pm$2.07 &  80.90$\pm$2.49 &  64.63$\pm$5.87 &  76.30$\pm$2.49 \\
CAIM \cite{george2023bridging} &  99.45$\pm$0.12 & 3.67$\pm$0.33 &  90.92$\pm$1.30 &  79.64$\pm$2.46 &  91.58$\pm$0.68 \\ \midrule 
\textbf{\ssmb} &  \textbf{99.70$\pm$0.08} & \textbf{2.59$\pm$0.28} &  \textbf{92.80$\pm$0.71} &  \textbf{84.04$\pm$1.71} &  \textbf{94.50$\pm$1.44} \\
\bottomrule
\end{tabular}

  }

\end{table*}

\textbf{Databases used:} To evaluate the performance of MAHFR we need to perform experiments on a dataset that contains images of subjects in multiple different modalities. We created new protocols in the MCXFace dataset to facilitate these evaluations. 

\textbf{MCXFace Dataset:} The MCXFace Dataset \cite{george2022prepended} includes images of 51 individuals taken in diverse lighting conditions across three sessions, featuring multiple channels such as RGB, thermal, near-infrared (850 nm), short-wave infrared (1300 nm), depth, stereo depth, and RGB-estimated depth. All channels are synchronized for spatial and temporal consistency. The dataset is divided into five folds per protocol, with subjects randomly divided into training or development groups. Annotations for eye centers are included for all images. Protocols follow the naming scheme $<SOURCE>-<TARGET>-split<split>$. The dataset is available at \footnote{\url{https://www.idiap.ch/dataset/mcxface}}.

\textit{Protocols: } The protocols shipped with the dataset include both homogeneous and heterogeneous experimental protocols. For example, $VIS-THERMAL-split4$ denotes a heterogeneous experiment with RGB (VIS) as the source for enrollment and THERMAL as the probe channel, aligning with the third split in the VIS-THERMAL five-fold partitions. We have introduced a new set of protocols, $VIS-UNIVERSAL-split<split>$, to benchmark MAHFR capabilities, where enrollment samples come from the VIS domain and probe samples can be any modality (Thermal, Near Infrared, or Shortwave Infrared). Each protocol features a training set with identities in both SOURCE and TARGET modalities and a development (dev) set with SOURCE images for enrollment and TARGET images for probing. Training and model selection are conducted within the training set, while the dev set scores are used solely for comparison, not training or model selection. For thorough evaluation, experiments should run across all five splits of the protocol, reporting mean values and standard deviations. We additionally compute the modality-wise performance separately to compare different methods.

\begin{table*}[htb!]
  \centering
  \caption{Experimental results on VIS-UNIVERSAL protocol of the MCXFace dataset--modality-wise performance breakdown.}
  \label{tab:mcxface_detailed}
  \resizebox{0.75\textwidth}{!}{
\begin{tabular}{ccccccc}
\toprule
\textbf{Model} & \textbf{Modality} &             \textbf{AUC} &           \textbf{EER} &              \textbf{Rank-1} &         \textbf{VR@FAR=0.1$\%$} &            \textbf{VR@FAR=1$\%$} \\
\midrule
DSU \cite{george2022prepended} &NIR & 100.00$\pm$0.00 &  0.00$\pm$0.00 & 100.00$\pm$0.00 & 100.00$\pm$0.00 & 100.00$\pm$0.00 \\
PDT \cite{george2022prepended}     &NIR & 99.89$\pm$0.15 &  0.98$\pm$1.34 &  97.72$\pm$3.58 & 91.23$\pm$12.48 &  96.48$\pm$4.87 \\
CAIM \cite{george2023bridging}     &NIR & 100.00$\pm$0.00 & 0.00$\pm$0.00 & 100.00$\pm$0.00 & 100.00$\pm$0.00 & \textbf{100.00$\pm$0.00} \\ 
\textbf{\ssmb}     &NIR & 100.00$\pm$0.00 & 0.00$\pm$0.00 & 100.00$\pm$0.00 & 100.00$\pm$0.00 & \textbf{100.00$\pm$0.00} \\ \midrule 
DSU \cite{george2022prepended}    &SWIR & 100.00$\pm$0.00 &  0.21$\pm$0.10 &  99.80$\pm$0.20 &  99.38$\pm$0.47 &  99.94$\pm$0.13 \\
PDT \cite{george2022prepended}    &SWIR & 99.35$\pm$0.89 &  2.57$\pm$3.31 & 90.61$\pm$12.82 & 83.25$\pm$22.46 & 91.66$\pm$11.42 \\
CAIM \cite{george2023bridging}    &SWIR &  99.99$\pm$0.02 & 0.27$\pm$0.25 &  99.60$\pm$0.48 &  98.83$\pm$1.40 &  99.89$\pm$0.25 \\ 
\textbf{\ssmb}    &SWIR & 100.00$\pm$0.00 & 0.18$\pm$0.11 &  99.86$\pm$0.27 &  99.67$\pm$0.26 & \textbf{100.00$\pm$0.00} \\ \midrule 
DSU \cite{george2022prepended} &THERMAL &  90.05$\pm$2.23 & 17.86$\pm$1.97 &  55.12$\pm$3.21 &  27.92$\pm$6.15 &  51.13$\pm$5.48 \\
PDT \cite{george2022prepended} &THERMAL & 91.38$\pm$3.80 & 16.07$\pm$4.14 &  56.55$\pm$9.41 & 30.33$\pm$10.34 & 52.74$\pm$11.01 \\
CAIM \cite{george2023bridging} &THERMAL &  98.11$\pm$0.39 & 6.95$\pm$0.60 &  74.51$\pm$3.45 &  38.86$\pm$9.70 &  70.82$\pm$4.16 \\ 
\textbf{\ssmb} &THERMAL &  98.89$\pm$0.27 & 5.08$\pm$0.79 &  79.41$\pm$2.28 &  43.74$\pm$6.51 &  \textbf{78.93$\pm$6.67} \\

\bottomrule
\end{tabular}

  }

\end{table*}

\textbf{HFR datasets:} In addition to performing experiments in the MCXFace dataset, we also evaluate the performance of the proposed approach in standard HFR benchmarks where only one source target modality pair is present. The Tufts Face Database \cite{panetta2018comprehensive}, features face images across various modalities, designed for face recognition tasks. Specifically, for assessing VIS-Thermal face recognition performance, we use the thermal images within this dataset, with 113 identities, following standard protocol in \cite{fu2021dvg}. The SCFace dataset \cite{grgic2011scface}, consists of high-quality images for enrollment, and lower-quality samples from surveillance cameras at varying distances as probes. We use the most challenging ``far'' protocol for reporting the results. The CUHK Face Sketch FERET Database (CUFSF), introduced by Zhang et al. \cite{zhang2011coupled}, consists of 1194 faces from the FERET dataset \cite{phillips1998feret}, each paired with an artist-drawn sketch. These sketches, often exaggerated, present a challenge for the face recognition task. We follow protocols in the previous works for comparison \cite{george2022prepended}.

\textbf{Baseline methods:} We compare the proposed approach against different methods in recent literature specifically the work on domain-specific units DSU \cite{de2018heterogeneous}, and its re-implementation using Iresnet100 \cite{george2022prepended}. Prepended domain transformer \cite{george2022prepended} and CAIM \cite{george2023bridging}. Since these methods were originally designed for a pair of modalities, we maintain the asymmetric nature for a fair comparison-- meaning, separate paths for the SOURCE and TARGET modalities. We use the same pre-trained face recognition model and weights in all these models and the proposed model.

\textbf{Metrics:} We evaluate the models based on several performance metrics frequently used in recent literature, such as Area Under the Curve (AUC), Equal Error Rate (EER), Rank-1 identification rate, and Verification Rate at specific false acceptance rates (0.01\%, 0.1\%, 1\%, and 5\%).

\subsection{Experimental results}

In this section, we compare the effectiveness of the proposed approach for MAHFR. We further perform evaluations with standard \hfr datasets to show the effectiveness of the proposed approach.

\textbf{MAHFR performance on MCXFace dataset:} We have performed experiments on the baseline methods and the proposed approach on the MCXFace dataset on the newly introduced $VIS-UNIVERSAL$ protocols. Here the aggregate performance over all the modalities is computed. The results of these protocols are shown in Tab. \ref{tab:mcxface_combined}. It can be seen that the proposed approach outperforms other methods by a large margin. For three different probe modalities combined (near-infrared, thermal, and shortwave infrared), the proposed approach achieves an average verification rate of 94.50 \%. 
\begin{table}[h]
  \centering
  \caption{Experimental results on VIS-UNIVERSAL protocol of the MCXFace dataset.}
  \label{tab:mcxface_ablation_expert}
  \resizebox{0.85\columnwidth}{!}{

\begin{tabular}{cccc}
\toprule
\textbf{Experts (N)} &             \textbf{AUC} &           \textbf{EER} &              \textbf{Rank-1}  \\
\midrule
1 &  99.66$\pm$0.13 & 2.85$\pm$0.50 &  91.89$\pm$1.17  \\
2 &  99.65$\pm$0.05 & 2.87$\pm$0.32 &  91.54$\pm$0.51  \\
3 &  99.64$\pm$0.10 & 2.82$\pm$0.53 &  91.54$\pm$1.43  \\
\textbf{4} &  \textbf{99.70$\pm$0.08} & \textbf{2.59$\pm$0.28} &  \textbf{92.80$\pm$0.71}  \\
5 &  99.68$\pm$0.07 & 2.75$\pm$0.41 &  92.03$\pm$1.74  \\

\bottomrule
\end{tabular}
  }

\end{table}
To analyze the performance further, we further separate the performance of the models in different modalities. The results are shown in Tab. \ref{tab:mcxface_detailed}. From this table, it can be seen that out of the modalities present, thermal modality is the most challenging one. The proposed approach significantly improves the performance in the thermal channel achieving an average verification accuracy of 78.9\% at an FAR (1\%). The proposed method achieves 100 \% verification accuracy at FAR 1\% operating point for both NIR and SWIR modalities. Overall, the proposed approach significantly boosts the unified face recognition performance in the MCXFace dataset.

\textbf{Effect of number of experts:} We have performed a set of experiments to evaluate the change in performance 
with the number of experts used. The results are shown in Tab. \ref{tab:mcxface_ablation_expert}, it can be seen that we achieve the best performance when the number of experts is four. Increasing the number of experts does not improve the performance further. It is to be noted that the experts in our framework is a single fully connected layer and the additional computational load required for more experts is minimal.

\textbf{Experiments with standard HFR datasets:} We have performed an extensive evaluation using three standard Heterogeneous Face Recognition (HFR) datasets to benchmark our method against previously proposed approaches in the literature. Specifically, the objective of these evaluations was to compare the performance of our proposed method in scenarios where only a single pair of source and target modalities are present. We have conducted experiments using both one and two experts and have reported the performance for both configurations. The Tufts Face dataset evaluates the performance in the visible-thermal heterogeneous scenario. As seen in the results presented in Table \ref{tab:tufts}, our method outperformed other methods in terms of the Verification rate (80.33\% at 1\% FAR) and achieved the best Rank-1 accuracy with two experts. SCFace dataset addresses heterogeneity in terms of image quality, we specifically focus on the `far` protocol which has the lowest quality images. The results in Table \ref{tab:scface}, show that our method attained the highest Rank-1 accuracy at 87.73\% and the second-best Equal Error Rate (EER) of 5.91\%. Here the best results are obtained with a single expert, this could be because the source and target modality are RGB and the difference is only in the image quality. CUFSF dataset evaluates the performance of heterogeneous face recognition in visible-to-sketch images, and from the results in Tab. \ref{tab:cufsf} it can be seen that the proposed approach achieves s Rank-1 accuracy of 81.67\% using two experts significantly outperforming state-of-the-art methods. Overall, the proposed method not only achieves competitive performance but often surpasses existing SOTA methods in the HFR dataset protocols, highlighting its effectiveness.

\begin{table}[h]
  \centering
  \caption{Experimental results on VIS-Thermal protocol of the Tufts Face dataset.}
  \label{tab:tufts}
  \resizebox{0.95\columnwidth}{!}{
  \begin{tabular}{lccc}
    \toprule
    \textbf{Method} & \textbf{Rank-1} & \textbf{VR@FAR=1$\%$} & \textbf{VR@FAR=0.1$\%$}  \\ \midrule
      LightCNN \cite{Wu2018ALC} & 29.4 & 23.0 & 5.3 \\
      DVG \cite{fu2019dual} & 56.1 & 44.3 & 17.1 \\
      DVG-Face \cite{fu2021dvg} & 75.7 & 68.5 & 36.5 \\ 
      DSU-Iresnet100 \cite{george2022prepended} & 49.7 & 49.8 & 28.3 \\   
      
      PDT \cite{george2022prepended} & 65.71 & 69.39 & 45.45 \\ 
      
      CAIM \cite{george2023bridging}  & 73.07 & 76.81 & 46.94 \\ \midrule

      \textbf{\ssmb} (N=1) & 75.04 & 78.29 & 53.99 \\
      \textbf{\ssmb} (N=2) & \textbf{78.46} & \textbf{80.33}   &\textbf{54.55} \\

      \bottomrule
  
  \end{tabular}
  }

\end{table}

\begin{table}[h]
  \caption{Performance of the proposed approach in the SCFace dataset. }
  \label{tab:scface}
  \centering
  \resizebox{0.98\columnwidth}{!}{%
  \begin{tabular}{lcrrrr}
  \toprule
  \textbf{Protocol}             & \textbf{Method} & \textbf{AUC}   & \textbf{EER}   & \textbf{Rank-1}    & \begin{tabular}[c]{@{}c@{}} \end{tabular} \\ \midrule

\multirow{4}{*}{Far}     
                                &DSU-Iresnet100 \cite{george2022prepended} & 97.18 & 8.37 & 79.53 \\

                             & PDT \cite{george2022prepended}   &  98.31 & 6.98 &  84.19            \\ 
                             &  CAIM \cite{george2023bridging} &   \textbf{98.81} &   \textbf{5.09} &   86.05  \\ \cmidrule{2-6}

                             &\textbf{\ssmb} (N=1) & 98.77 &  5.91 &  \textbf{87.73} \\
                             &\textbf{\ssmb} (N=2) & 98.67 & 6.36 & 86.82 \\

  \bottomrule
  \end{tabular}
  }

  \end{table}

\section{Discussions}
\label{sec:discussions}
The proposed framework introduces a new way of performing cross-domain face recognition in a modality-agnostic manner. The model trained using our framework can produce embeddings for any modality used in the training. The embeddings obtained share the same embedding space and can be used for heterogeneous or homogeneous matching. To enable this, we utilize a training paradigm that leverages supervision from a teacher network -- enabling stable training without overfitting. The addition of multiple modalities at the training time with the adaptive routing enables the network to dynamically adapt the parameters depending on the input feature maps. It is important to note that specific modality labels are not required at inference time, meaning any face image can be used as the input to compute the representation of face embedding, and the network routes the data automatically to produce the embedding for the specific modality to maximize the matching performance. This could be useful in scenarios where specific source target modality pairs do not need to be trained separately and greatly simplifies the HFR process. It is worth noting that the \textit{StyleExperts} in our framework is lightweight, and the switch routing mechanism can be optimized further to reduce computation at inference time.

\begin{table}[ht]
\caption{CUFSF: Rank-1 recognition rate in sketch to photo recognition}
\label{tab:cufsf}

\begin{center}
\resizebox{0.45\columnwidth}{!}{%
  \begin{tabular}{lrr}
    \toprule
    \textbf{Method} & \textbf{Rank-1} \\ \midrule
    IACycleGAN \cite{fang2020identity} &64.94 \\
    DSU-Iresnet100 \cite{george2022prepended} & 67.06 \\ 
    PDT \cite{george2022prepended}  & 71.08 \\ 
    CAIM \cite{george2023bridging} & 76.38 \\ \midrule
    \textbf{\ssmb} (N=1) &  81.14\\
    \textbf{\ssmb} (N=2) & \textbf{81.67} \\
    \bottomrule 
  \end{tabular}

  }
\end{center}
\vspace{-5mm}
\end{table}

\section{Conclusion}
\label{sec:conclusion}

In this work, we introduce a novel modality-agnostic framework capable of performing cross-domain face recognition without the need for explicit modality labels. This eliminates the need for training separate models for each source-target pairing, allowing for the adoption of a single, unified model. We achieve this by introducing trainable Switch Style Modulation Blocks (SSMB) to dynamically route input feature maps, effectively reducing the domain gap with minimal computational overhead, thereby making our solution both efficient and scalable. Our proposed approach thus facilitates mapping various face modalities into a shared embedding space. Experiments on the MCXFace dataset and standard HFR benchmarks demonstrate the effectiveness of our method. The source code and protocols will be made available publicly.
\section*{Acknowledgements}
\label{sec:ack}
This research is based upon work supported in part by the Office of the Director of National Intelligence (ODNI), Intelligence Advanced Research Projects Activity (IARPA), via [2022-21102100007]. The views and conclusions contained herein are those of the authors and should not be interpreted as necessarily representing the official policies, either expressed or implied, of ODNI, IARPA, or the U.S. Government. The U.S. Government is authorized to reproduce and distribute reprints for governmental purposes notwithstanding any copyright annotation therein.

{\small
\bibliographystyle{ieee}
\bibliography{egbib}
}

\end{document}